\documentclass[letterpaper]{article} 
\usepackage{aaai2026}  
\usepackage{times}  
\usepackage{helvet}  
\usepackage{courier}  
\usepackage[hyphens]{url}  
\usepackage{graphicx} 
\usepackage{natbib}  
\usepackage{caption} 
\usepackage{algorithm}
\usepackage{algorithmic}

\usepackage{amsmath}
\usepackage{cleveref}
\usepackage{booktabs}
\usepackage{textcomp}
\usepackage{utfsym}
\usepackage{multirow}
\usepackage{xspace}
\usepackage{pifont}
\usepackage{amssymb}
\usepackage{enumitem}
\usepackage{times}
\usepackage{helvet}
\usepackage{courier}
\usepackage{xcolor}
\usepackage{amssymb}
\usepackage[misc]{ifsym}
\usepackage{newfloat}
\usepackage{listings}
\DeclareCaptionStyle{ruled}{labelfont=normalfont,labelsep=colon,strut=off} 
\floatstyle{ruled}
\newfloat{listing}{tb}{lst}{}
\floatname{listing}{Listing}
\newcommand{\ourdataset}{Cut-VOS\xspace}
\newcommand{\numvideos}{100\xspace}
\newcommand{\numobjects}{174\xspace}
\newcommand{\nummasks}{10.2K\xspace}
\newcommand{\numcategories}{11\xspace}
\newcommand{\numshots}{648\xspace}
\newcommand{\ourmethod}{SAAS\xspace}

\newcommand{\eg}{\emph{e.g}}
\newcommand{\etc}{\emph{etc}}

\newcommand{\jf}{$\mathcal{J} \& \mathcal{F}$\xspace}
\newcommand{\jt}{$\mathcal{J}_t$\xspace}
\newcommand{\xmarkg}{\textcolor{lightgray}{\ding{55}}\xspace}

\newcommand{\myrule}{\specialrule{0.08em}{.05em}{.05em}}

\title{Segment Anything Across Shots: A Method and Benchmark}
\author{
    Hengrui Hu, Kaining Ying, Henghui Ding\thanks{Corresponding author (hhding@fudan.edu.cn).} 
}
\affiliations{
    Fudan University\\
    \url{https://henghuiding.com/SAAS/}
}
\begin{document}

\maketitle

\begin{abstract}

This work focuses on multi-shot semi-supervised video object segmentation (MVOS), which aims at segmenting the target object indicated by an initial mask throughout a video with multiple shots. The existing VOS methods mainly focus on single-shot videos and struggle with shot discontinuities, thereby limiting their real-world applicability. 
We propose a transition mimicking data augmentation strategy (TMA) which enables cross-shot generalization with single-shot data to alleviate the severe annotated multi-shot data sparsity, and the Segment Anything Across Shots (\ourmethod) model, which can detect and comprehend shot transitions effectively.
To support evaluation and future study in MVOS, we introduce \ourdataset, a new MVOS benchmark with dense mask annotations, diverse object categories, and high-frequency transitions. Extensive experiments on YouMVOS and \ourdataset demonstrate that the proposed \ourmethod achieves state-of-the-art performance by effectively mimicking, understanding, and segmenting across complex transitions. The code and datasets are released at \url{https://henghuiding.com/SAAS/}.

\end{abstract}

\section{Introduction} \label{sec:intro}

Semi-supervised video object segmentation (VOS)~\cite{2017OSVOS} aims to segment and track the target object throughout a video sequence, given its mask in the first frame as a prompt. This task has received increasing attention~\cite{2024SAM2} in the research community because of its broad applicability in human–robot interaction, video editing, autonomous driving, and annotation assistance, \etc.

\begin{figure*}[t]
    \centering
    \includegraphics[width=0.987\textwidth]{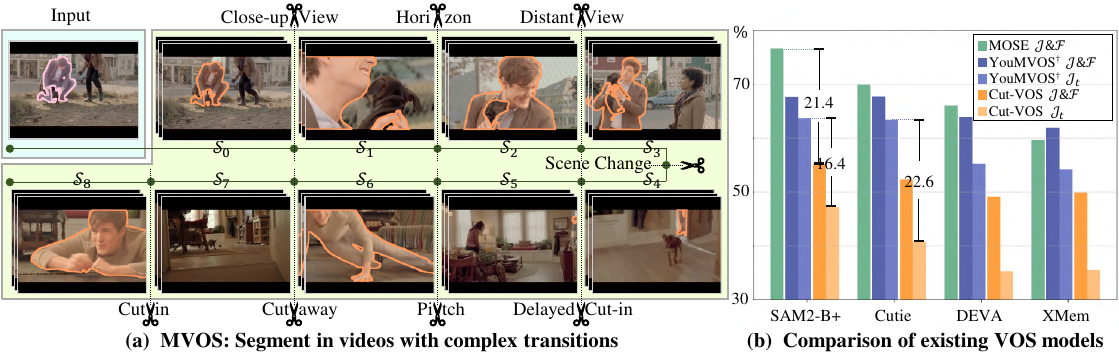}
    \caption{
    This work focuses on an underexplored task of multi-shot video object segmentation (MVOS).
    As shown in (a), the significant variations in object appearance, spatial location, and background across shots pose major challenges in MVOS. We introduce Cut-VOS, a challenging MVOS benchmark with high transition diversity to support this task. As shown in (b), on Cut-VOS, SAM2-B+ exhibits a 21.4\% \jf drop compared to the challenging single-shot MOSE dataset and a 16.4\% \jt drop compared to YouMVOS$^\dagger$, a sampled MVOS dataset YouMVOS annotated by our team strictly following its original protocol. The metric \jt specifically measures cross-shot segmentation performance, further highlighting the difficulty of Cut-VOS.
    }
    \label{fig:teaser}
\end{figure*}

Despite notable progress, existing VOS methods predominantly focus on single-shot videos, overlooking the increasing prevalence of multi-shot videos~(see~\Cref{fig:teaser}~(a)) in real-world Internet content. This oversight on \textbf{multi-shot video object segmentation}~(\textbf{MVOS}) has led to a widening gap between academic research and practical deployment.
The current representative VOS methods, \eg. XMem~\cite{2022Xmem}, DEVA~\cite{cheng2023tracking}, Cutie~\cite{2024Cutie}, and SAM2~\cite{2024SAM2} exhibit a notable performance degradation when exposed to complex shot transitions. As shown in~\Cref{fig:teaser}~(b), SAM2-B+ suffers a 21.4\% \jf drop on the MVOS benchmark compared to MOSE~\cite{2023MOSE}, highlighting their limitations in the applications of edited videos, multi-camera systems, and high-mobility platforms.

\begin{figure}
    \centering
    \includegraphics[width=0.998\linewidth]{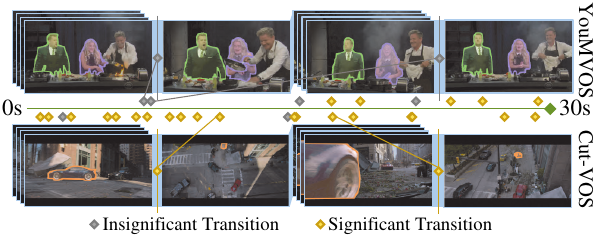}
    \caption{
    The comparison between YouMVOS and our proposed Cut-VOS benchmark. Cut-VOS is distinguished from YouMVOS by frequent, significant transitions and more variety in complex scenarios.
    }
    \label{fig:fig_dataset_comp}
\end{figure}

To our knowledge, YouMVOS~\cite{2022YouMVOS} is currently the only dataset that supports MVOS. 
However, upon reviewing the playlists provided in their dataset, we find that the dataset falls short in fully reflecting the challenges of MVOS task. Specifically, the dataset contains only sparse shot transitions, exhibits a limited diversity of object categories with a predominant focus on humans, and lacks screening or categorization of transition types, as shown in~\Cref{fig:fig_dataset_comp}.  Furthermore, the mask annotations of YouMVOS have not been open-sourced to date, making it unavailable for subsequent model development and training.

To address the lack of multi-shot training data, we propose the \textbf{T}ransition \textbf{M}imicking Data \textbf{A}ugmentation (\textbf{TMA}) strategy, which simulates diversiform shot transitions on single-shot datasets to enable effective multi-shot segmentation training without relying on native multi-shot annotations. 
Meanwhile, the deficiencies of previous methods in complex multi-shot videos, as shown in \Cref{fig:teaser}~(b), prompt us to develop a specialized cross-shot segmentation method,
\textbf{S}egment \textbf{A}nything \textbf{A}cross-\textbf{S}hot (\textbf{\ourmethod}), equipped with transition detection and comprehension modules. These modules jointly detect and interpret shot transitions using adjacent frames along with background context, guided by two auxiliary training objectives. Additionally, we introduce a training-free memory refinement mechanism through a local memory bank that stores fine-grained object features to enhance segmentation quality across transitions.

To fairly evaluate cross-shot segmentation performance and better reflect the complexity of real-world multi-shot videos, 
we introduce a new MVOS benchmark, \textbf{C}omplex M\textbf{u}lti-sho\textbf{t} \textbf{V}ideo \textbf{O}bject \textbf{S}egmentation (\textbf{\ourdataset}), containing \nummasks instance masks for \numobjects unique objects in \numvideos videos. Compared to YouMVOS, the proposed \ourdataset provides 1.6$\times$ higher shot transition frequency and 3$\times$ more object categories. The transition types are manually screened to ensure greater diversity and difficulty. For qualitative comparison, we build YouMVOS$^\dagger$\footnote{All experiments on YouMVOS in this paper are conducted on YouMVOS$^\dagger$, a manually annotated version constructed by us, as the original dataset does not release mask annotations.} test split by sampling and annotating 30 videos across 10 genres from the playlist, strictly following their announced protocol. Compared to YouMVOS,
the models perform significantly worse on \ourdataset, as shown in~\Cref{fig:teaser} (b), indicating a substantial difficulty gap. 
Extensive experiments demonstrate that \ourmethod achieves consistent improvements across both YouMVOS and \ourdataset.

Overall, the key contributions of this work are as follows:

\begin{itemize}
\item We introduce a new VOS training strategy, \textbf{T}ransition \textbf{M}imicking Data \textbf{A}ugmentation (TMA), to alleviate data sparsity by simulating shot transitions, thereby promoting the model's multi-shot segmentation capacity using only single-shot datasets.
\item To the best of our knowledge, the proposed \textbf{\ourmethod} is the first semi-supervised VOS method specialized for multi-shot videos. It incorporates online transition detection, transition comprehension, and local visual cue encoding. Extensive experiments demonstrate its robustness and effectiveness in complex multi-shot scenarios.
\item To facilitate future research in MVOS, we introduce \textbf{C}omplex M\textbf{u}lti-sho\textbf{t} \textbf{V}ideo \textbf{O}bject \textbf{S}egmentation (\textbf{\ourdataset}) dataset, which will become the first fully open-sourced MVOS benchmark with mask annotations upon publication. \ourdataset provides diverse object categories and carefully curated transition types to evaluate cross-shot tracking performance. 
\end{itemize}

\section{Related Work} \label{sec:rel_works}

\begin{figure*}[t!]
    \centering
    \includegraphics[width=\textwidth]{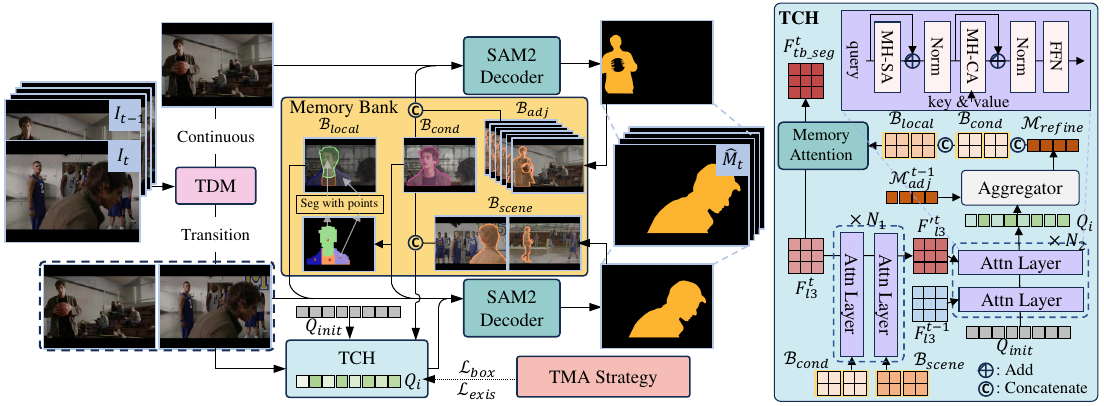}
    \caption{
     The overall pipeline of our proposed Segment Anything Across Shots (\ourmethod) method, consisting of three new components, Transition Detection Module (TDM), Transition Comprehension Module (TCH), and local memory bank $\mathcal{B}_{local}$. Transition Mimicking Augmentation (TMA) is employed to train the model by synthesizing high-quality multi-shot training samples using annotated single-shot videos.}
    \label{fig:method}
\end{figure*}

\textbf{Video Object Segmentation.} Video object segmentation (VOS)~\cite{ding2023mevis,ding2025mosev2,ding2025mevis,ding2025multimodal,2017OSVOS,ying2025move,lin2019agss, huang2020fast,he2025survey,liu2025lsvos} aims at tracking and segmenting the objects in a video sequence, given the mask in the first frame. Early methods~\cite{2018monet, perazzi2017learning} are mostly fine-tuning-based. They model inter-frame correlations via fine-tuning during inference. Matching-based methods~\cite{2018fast, 2019capsulevos, 2021sstvos} generate an object prototype embedding from the conditional frame, performing pixel-level matching to classify each pixel as foreground or background. Propagation-based methods~\cite{2018reinforcement, hu2018motion, 2020space, wang2019ranet} leverage the previous frames and predictions to guide the segmentation on the current frame. For better use of historical information, recent methods introduce a memory bank to compress and store previous frames. For example, XMem~\cite{2022Xmem} conducts multiple granularities of memories, while Cutie~\cite{2024Cutie} enriches the memory bank with object-specific queries. Most recently, SAM2~\cite{2024SAM2} extends SAM~\cite{2023segment} to the video domain, yielding a remarkable improvement via a robust memory architecture and large-scale training. However, these previous methods only focus on single-shot videos, lacking solid cross-shot tracking capacity, which leads to their limited applications. This work aims to generalize VOS to multi-shot videos, bridging the gap between the current research and practical requirements.

\textbf{Multi-shot Video Understanding.} Multi-shot videos, which circulate on the internet at an increasingly large scale, have gradually attracted the attention of the computer vision community. Most early works~\cite{1986computational, 2004automatic, 2006effective} aim to detect the shot boundaries with manual features. With the development of deep learning, some methods~\cite{2017DSDB, 2024transnetv2, bouyahi2020video, wang2021shot} adapt 3D-CNN~\cite{20123d} and dilated filter~\cite{chen2017deeplab,2017dilated} to improve model accuracy. Meanwhile, some works collect multi-shot videos in their video captioning benchmarks~\cite{2016msr, 2017dense, 2018towards}, asking the model to generate video descriptions. Recently, Shot2Story~\cite{Story20K} and MMBench-Video~\cite{2024MultiShotUnd} posed more fine-grained questions, requiring clip-wise understandings to answer. MUSES~\cite{MultiShotTAL} focuses on the multi-shot temporal event localization task which requests dense frame labels. However, these works still lack the exploration of pixel-level instance segmentations~\cite{isda,ctvis}. This paper specifically targets fine-grained segmentation in multi-shot videos.

\section{Methodology} \label{sec:method}
\subsection{Overview}
\Cref{fig:method} shows an overview of our approach, which contains the proposed  Transition Mimicking  Data Augmentation (TMA) training strategy and a transition-aware method, Segment Anything Across Shots (\ourmethod), built upon the SAM2 to generalize VOS to multi-shot videos. Given a video $\mathcal{V}=\{ I_t \}^T_{t=1}$ with $T$ frames, and the first frame $I_0$ with ground truth mask $M_0$, \ourmethod firstly applies SAM2 image encoder to extract multi-level visual features $\{ F^t_{li} \}_{i=1,2,3}$. At each timestep $t$, SAAS introduces the Transition Detection Module~(TDM) to detect if a shot transition occurs and subsequently directs to diverse segmentation strategies. For the detected transitions, the following Transition Comprehension Module~(TCH) further comprehends them, generates compressed transition state representation $Q_i$, thereby refining previous memories. To capture the local fine-grained features of objects, we also propose the local memory bank $B_{local}$, to partition the target and store corresponding information unsupervisedly. The conditional memories from $B_{cond}$ and features stored in $B_{local}$ are then concatenated to generate the features prepared to be segmented $F^t_{tb\_seg}$, used to finally predict $\hat{M}_t$ by the mask decoder. The entire architecture is trained via the TMA strategy, with two additional objectives.

\subsection{Transition Mimicking Augmentation}
One of the most critical challenges for MVOS is the lack of available training data. To address this issue, we propose \textbf{T}ransition \textbf{M}imicking Data \textbf{A}ugmentation (\textbf{TMA}), a new strategy which synthesizes quality-approved multi-shot training samples from annotated single-shot videos by simulating diverse transitions. TMA enables the effective MVOS training utilizing existing single-shot VOS datasets, significantly alleviating data scarcity. 

\begin{figure}[t]
    \centering
    \includegraphics[width=0.995\linewidth]{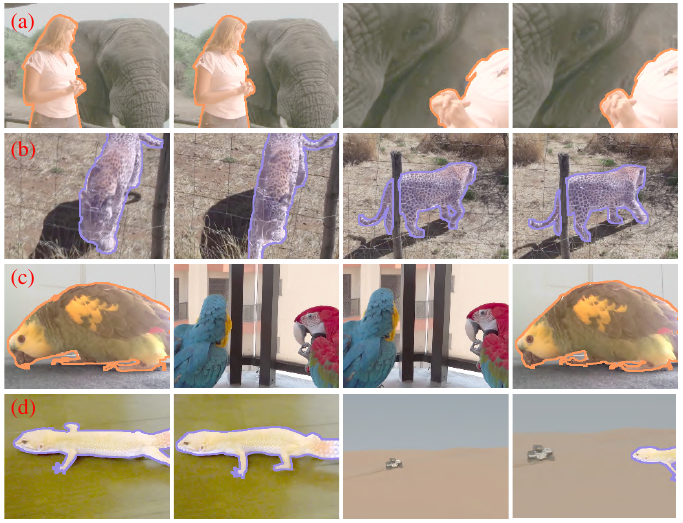}
    \caption{
    Some visualization cases of our proposed TMA strategy. (a) Random strong transforms. (b) Single transition across different temporal segments from the same video. (c) Multiple transitions, conducting a case with \textit{cut in} and \textit{cut away}. (d) Single transition to another video, with random replication and gradual translations.  
    }
    \label{fig:TMA}
\end{figure}

We show some primary patterns involved in TMA in~\Cref{fig:TMA}. TMA maintains a conventional 8-frame continuous sampling strategy in previous VOS works with a probability $1 - p_{trans}$, 
otherwise performs a transition mimicking operation. Specifically, TMA conducts a single transition (as shown in (a), (b), and (d)) with a probability $p_{once}$, otherwise applies multiple transitions (as depicted in (c)). 
For each expected transition, TMA employs a well-defined framework with several control random variables to generate different transition patterns.
For example, case (a) retains a continuous 8-frame sampling but applies strong transformations, including horizon flipping, random scaling, and random affine on posterior frames after the transition. This pattern simulates common view transitions, like \textit{close-up view} or \textit{distant view}. Case (b) cuts to a different segment from the same video, with a higher probability of sampling more further frames. The substantial temporal gap between the two clips often results in significant changes in object poses and camera viewpoints. Case (c) cuts to an unrelated video and cuts back later, like the \textit{cut away} and \textit{cut in} transitions. Case (d) cuts to an unrelated video while replicating the object with a random, gradual translation, simulating the \textit{scene change} and the \textit{delayed cut in} transitions effectively. 
TMA fully combines these patterns to preserve data richness while carefully avoiding ambiguous samples and anomalous noises. More details are offered in the appendix.

\subsection{Transition Detection and Comprehension}

\textbf{Transition Detection Module.} SAAS employs a light-weight \textbf{T}ransition \textbf{D}etection \textbf{M}odule (\textbf{TDM}) to detect different shot segments and occurring transitions in video sequences. Inspired by previous shot boundary detection methods~\cite{2017transnet, 2024transnetv2}, we conduct a dilated convolution pyramid~\cite{chen2018encoder, chen2019rethinking} as TDM. At each timestep $t$, TDM predicts a probability score for current frame $I_t$, directing to different pipelines:
\begin{equation}\label{eq:TDM}
\hat{p}_{i,tr}=\text{Sigmoid}(\mathcal{F}_{\text{TDM}}(F^t, F^{t-i}_{i=1,2,...,N})),
\end{equation}
where $\mathcal{F}_{\text{TDM}}$ indicates the main network of TDM which uses the adjacent $N$ frames for detection. When $\hat{p}_{i,tr} < \tau_{tr}$, SAAS passes through the SAM2 segmentation pipeline (the upper part in~\Cref{fig:method}) directly, and only encodes the memory $\mathcal{M}_t$ into the bank $\mathcal{B}_{adj}$. Otherwise, SAAS recognizes the transition occurs and adopts a transition segmentation strategy instead (the down part). Extracted features ${F^t}$ and $F^{t-1}$, along with few memory banks, are fed to the TCH. TCH compresses them to refine the memory tokens, followed by the segmentation head to achieve a cross-shot segmentation. Meanwhile, the memory $\mathcal{M}_t$ is encoded and stored in a special memory bank $\mathcal{B}_{scene}$ instead, used to establish a necessary spatial scene understanding in TCH. 

\textbf{Transition Comprehension Module.} SAAS builds a \textbf{T}ransition \textbf{C}ompre\textbf{h}ension Module (\textbf{TCH}) to firstly associate stored scene information and then integrate adjacent frames to fully comprehend the occurring transition. 
Specifically, TCH reads out the background scene information from the banks $\mathcal{B}_{cond}$ and $\mathcal{B}_{scene}$. $\mathcal{B}_{scene}$ stores representative memories for the most closed $N_s$ shots. These memories are used to build an entire scene understanding, subsequently integrated into $F^t_{l3}$ via stacked attention layers, attaining $F'^t_{l3}$. Then, a trainable vector $Q_{init}$ passes through the module,  sufficiently interacts with the features of the previous frame and the current frame to comprehend the current transition:
\begin{equation}\label{eq:TCH}
Q_i^n = \text{Attn}(\text{Attn}(Q_i^{n-1}, F'^t_{l3}), F^{t-1}_{l3})),
\end{equation}
where $Q_i^0 = Q_{init}$, $n=\{1,2,...,N_2\}$. Attn represents a standard attention layer~\cite{vaswani2017attention}, consisting of a multi-head cross-attention, a multi-head self-attention, and a feed forward layer following previous ViT works~\cite {dosovitskiy2020image, liu2021swin} with a RoPE positional encoding~\cite {su2024roformer}.
To validate the process of transition state modeling, we incorporate two additional auxiliary objectives: presence prediction and bounding box regression. Presence prediction requires the model to predict the presence of the object on the next frame from the transition state representation $Q_i$, supervised by a BCE loss $\mathcal{L}_{exis}$. For the bounding box regression objective, the model learns a mapping from the previous bounding box and $Q_i$ to the post-transition bounding box, adopting a MCE loss $\mathcal{L}_{box}$. Simple MLP stacks suffice for these objectives.

Subsequently, an attention-based aggregator is introduced to decode the transition state $Q_i$ to refine the previous memory $\mathcal{M}^{t-1}_{adj}$. This decoding strategy ensures seamless compatibility with SAM2's well-trained segmentation head. The final refined memories are concatenated with memories from $\mathcal{B}_{cond}$ and $\mathcal{B}_{local}$ and fed to SAM2's memory attention module to prepare the features to be segmented $F^t_{tb\_seg}$.

\subsection{Local Memory Bank}
\label{LMB}

\begin{table*}[t!]
  \centering
  \normalsize
  \begin{tabular}{lccccccc}
  \myrule
      \textbf{Dataset} & \textbf{\#Videos} & \textbf{\#Objects} & \textbf{\#Masks} & \textbf{\#Shots} & \textbf{Trans. Frequency} & \textbf{Obj. Categories} & \textbf{Available} \\ 
      \hline
      YouMVOS & 200 & 492 & 431.0K \ \ & 13.4K & 0.222/s* & \ \ 4 & \xmarkg \\   
      YouMVOS-test & \ \ 30 & \ \ 78 &\ \ 64.6K* & \ \ 2.4K & 0.222/s* & \ \ 4 & \xmarkg \\ 
      \hline
       \textbf{\ourdataset (ours)}  &  \numvideos & \numobjects &\ \ \nummasks \ \ & \ \ \ \numshots & 0.346/s \ \ & \numcategories & \usym{2713}  \\  
  \myrule
  \end{tabular}
  \caption{The basic statistics for the Cut-VOS benchmark. * denotes the number is estimated via the corresponding description in the paper. Cut-VOS has 1.6× higher transition frequency and 3× more categories than the YouMVOS test split.}
  \label{tab:dataset_statistics}
\end{table*}

In a significant proportion of transitions, local object details can serve as critical segmentation cues, like the clothing of a person or the painted markings on a vehicle.
Previous VOS methods struggle to actively capture and recognize such fine-grained features.
Informed by such an observation, SAAS introduces a local memory bank $\mathcal{B}_{local}$ to capture and store the target's local details. Inspired by previous works~\cite{2019learnable, 2022tree, 2024gtms}, SAAS constructs a minimum spanning tree (MST) on the masked deepest feature map of the conditional frame $M_0\odot F^0_{l3}$ to simultaneously preserve semantic clustering and spatial structural information. By pruning low-weight edges in the tree, the target is unsupervisedly partitioned into multiple semantically coherent sub-regions on a low-resolution map. SAAS further adopts the center point of each partition as a positive point prompt, the rest as the negative to segment these sub-regions and extract corresponding fine-grained features at a high resolution. These detailed features are compressed as complementary object pointers and preserved in the local memory bank $\mathcal{B}_{local}$, which is leveraged to guide the cross-shot segmentation when a transition is detected. Notably, we set a proportion threshold $\tau_{p}$  (2.5\% in a common setting) to filter out too small objects, preventing over-partitioning them. 

\section{Cut-VOS Benchmark} \label{sec:dataset}
\subsection{Video Collection and Annotation} \label{sec:collection}

\begin{figure}
    \centering
    \includegraphics[height=4.3cm]{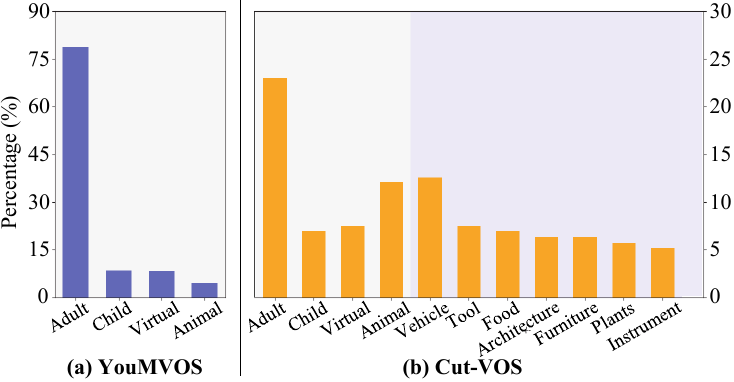}
    \caption{
    Comparison of object categories. \ourdataset contains 4 categories in YouMVOS and 7 new categories.
    }
    \label{fig:category_comp}
\end{figure}

The new challenging multi-shot video object segmentation~(MVOS) benchmark, \underline{C}omplex M\underline{u}lti-sho\underline{t} \underline{V}ideo \underline{O}bject \underline{S}egmentation~(Cut-VOS), collects large amounts of high-quality multi-shot videos from mainstream community media. 
The videos and objects are carefully selected to ensure the data samples are unambiguous.
The detailed object and transition distributions are shown in~\Cref{fig:category_comp} and~\Cref{fig:transition_comp}.

For mask annotation, our research team organizes and trains a cohort of highly responsible annotators and validators, establishing a robust annotation pipeline. Each annotated video undergoes a dual-review verification to ensure annotation quality assurance. For videos that are discovered with uncertain object correlations, we reconvened discussions to determine whether to keep or filter them.

\subsection{Dataset Statistics} \label{sec:statistics}
Overall, \ourdataset contains \numvideos videos, \numobjects annotated objects, and \nummasks high-quality masks, as shown in \Cref{tab:dataset_statistics}.
Cut-VOS outperforms the existing YouMVOS-test in three main aspects: 1) More videos and objects representing more diverse scenarios. 2) Carefully screened, multiple types of transitions with a 1.6 times higher frequency reaching 0.346/s, which makes the Cut-VOS more challenging. 3) 11 diversiform object categories which cover the YouMVOS as depicted in~\Cref{fig:category_comp}, containing 62\% actors and 38\% static objects. These characteristics make the Cut-VOS benchmark more complex and better aligned with real-world scenarios.

\subsection{Transition Analysis} \label{sec:analysis}

\begin{figure}
    \centering
    \includegraphics[height=4.3cm]{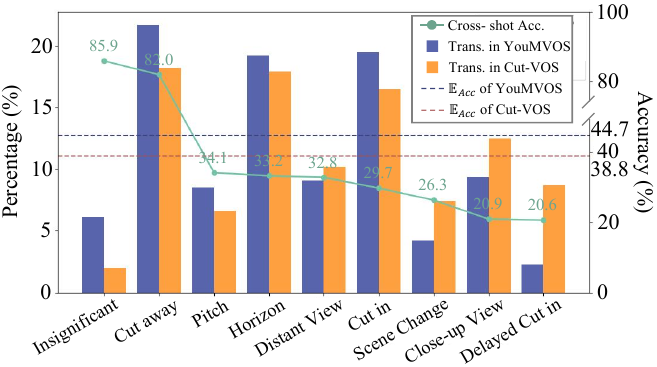}
    \caption{
    The average accuracies of different transition types on the SAM2-B+ model and their distribution across two benchmarks. The drop in expected accuracy shows \ourdataset's more challenging nature.
    }
    \label{fig:transition_comp}
\end{figure}

To better analyze the latent challenges in the MVOS task, we classify all shot transitions into 9 different categories:
\textit{cut in, cut away, delayed cut in} as existence types, 
and \textit{close up view, distant view, pitch transformation, horizon transformation, scene change, insignificance} as view types.
In specific cases, we allow the coexistence of an existence type and a view type in one transition.
Please refer to the appendix for detailed explanations and visualized examples.

\begin{table*}[t!]
  \centering
  \small
  \begin{tabular}{lccccccccccc}
  \myrule
  \multirow{2}{*}{Method} & \multirow{2}{*}{Venue} & \multirow{2}{*}{Param.(M)} & \multirow{2}{*}{FPS} & \multicolumn{4}{c}{YouMVOS} & \multicolumn{4}{c}{Cut-VOS} \\
  \cmidrule(lr){5-8} \cmidrule(lr){9-12}
        &    &    &    & $\mathcal{J}$ & $\mathcal{F}$ & \jf & \jt & $\mathcal{J}$ & $\mathcal{F}$ & \jf & \jt \\
  \hline
    XMem    & ECCV'22   & \ \ 62.2  & \textbf{45}    & 61.7  & 62.1  & 61.9  & 54.2  & 48.4  & 51.4  & 49.9  & 35.5  \\
    DEVA    & ICCV'23   & \ \ \underline{61.2}  & 37    & 63.3  & 64.5  & 63.9  & 55.2  & 47.3  & 50.8  & 49.1  & 35.3      \\
    Cutie   & CVPR'24   & \ \ \textbf{35.0}  & \underline{40}    & 67.3  & 68.1  & 67.7  & 63.4  & 51.0  & 53.6  & 52.3  & 40.8      \\
    Cutie$^\star$  & CVPR'24   & \ \ \textbf{35.0}  &  \underline{40}   & 67.9  & 68.8  & 68.4  & 64.7  & 50.0  & 52.7  & 51.4  & 40.0     \\
    SAM2-B+        & ICLR'25   & \ \ 80.9  &  22   & 67.6  & 67.6  & 67.6  & 63.7  & 54.0  & 56.4  & 55.2  & 47.2      \\
    SAM2-L         & ICLR'25   & 224.0 &  15   & 69.9  & 70.3  & 70.1  & 68.5  & 58.3  & 60.6  & 59.4  & 50.7      \\
    SAM2-B+$^\star$& ICLR'25   & \ \ 80.9  &  22   & 68.7  & 69.1  & 68.9  & 64.1  & 53.9  & 55.9  & 54.9  & 46.8      \\
    SAM2-L$^\star$ & ICLR'25   & 224.0 &  15   & 69.7  & 70.7  & 70.2  & 68.4  & 57.6  & 60.3  & 58.9  & 50.4     \\
  \hline
    Cutie+TMA       & -     & \ \ \textbf{35.0}  &  \underline{40}   & 69.1 & 70.0 &  69.6 &  65.4 & 52.0 & 55.0  & 53.5  & 43.1   \\
    \textbf{\ourmethod-B+~(Ours})   & AAAI'26 & \ \ 92.5  &  21   & \underline{73.4}  & \underline{73.7}  &  \underline{73.5}  & \underline{68.9}  & \underline{59.4} & \underline{61.9}  &  \underline{60.7}  & \underline{53.1}      \\
    \textbf{\ourmethod-L~(Ours})    & AAAI'26 & 235.6 & 14    & \textbf{74.0}  & \textbf{74.4}  & \textbf{74.2}  & \textbf{69.6}  & \textbf{60.5} & \textbf{63.6} &  \textbf{62.0} &  \textbf{54.0}       \\
  \myrule
  \end{tabular}
  \caption{Main results on YouMVOS and Cut-VOS benchmarks. $^\star$ denotes the model is directly trained on the YTVOS dataset without extra data augmentation. Bold and underlined indicate the best and the second-best performance in the tested methods.}
  \label{tab:main_results}
\end{table*}

We test the tracking accuracy on different transition types with the SAM2-B+ model to pinpoint existing bottlenecks.
As shown in \Cref{fig:transition_comp}, SAM2 performs well on \textit{cut away} and \textit{insignificance}, shows moderate competencies on \textit{pitch} and \textit{horizon} types, but drops ruinously on \textit{delayed cut-in}, \textit{close-up view}, and \textit{scene change} types (lower than 27\%).
The observation indicates that previous methods can recognize the object disappearing, but struggle with matching targets with abrupt visual appearance and absolute position shifts.
Cut-VOS filters out simple \textit{insignificance} and long duration \textit{cut away}, involving more difficult transitions to make the benchmark more challenging.
Compared to YouMVOS, the significant decrease of $\mathbb{E}_{Acc}$ (44.7\% to 38.8\%) reflects the challenges brought by screened complex transitions.

\section{Experiments}
\label{sec:experiments}

\begin{table}[t!]
  \centering
  \small
  \begin{tabular}{l|ccc|cc}
  \myrule
      ID & $\mathcal{B}_{local}$ & TMA & TCH & \jf & \jt \\ 
      \hline
      I   & \xmarkg &  \xmarkg  & \xmarkg & 55.2 & 47.2  \\   
      II        & \usym{2713} & \xmarkg & \xmarkg & 57.6 & 49.4   \\
      III       & \xmarkg & \usym{2713} & \xmarkg & 58.0  & 50.7   \\
      IV      & \usym{2713} & \usym{2713} & \xmarkg & 58.8 & 52.0  \\
      V       & \xmarkg & \usym{2713} & \usym{2713} & \underline{60.1}  & \underline{52.8}   \\
      VI   & \usym{2713} & \usym{2713} & \usym{2713} & \textbf{60.7}  & \textbf{53.1}  \\
  \myrule
  \end{tabular}
  \caption{The ablation study on different modules.}
  \label{tab:ablation_studies}
\end{table}

\textbf{Benchmark Setting.} We benchmark the proposed
\ourmethod and existing methods on Cut-VOS and YouMVOS under the semi-supervised VOS setting. Following previous works~\cite{2023MOSE,omniavs}, we compute \jf to quantify the region similarity and the contour accuracy of predictions. Besides, we additionally measure the cross-shot tracking capacity by computing region similarity \jt on post-transition frames. Given the ground truth shot set $\mathcal{S}$, for each shot $\mathcal{S}_i$ we calculate intersection over union~(IoU) on the first frame
$I_{tr}^i$ and the frame where the object firstly appears $I_{app}^i$ (defined as the first frame too if the object isn't present in the shot) separately, to accommodate different existence transitions, especially \textit{delayed cut in}. Then \jt is defined as:
\begin{equation}
\label{eq:Jt}
\mathcal{J}_t = \frac{1}{|\mathcal{S}|} \sum\limits_{i \in |\mathcal{S}|} \frac{\text{IoU}(\hat{M}_{tr}^i, M_{tr}^i) + \text{IoU}(\hat{M}_{app}^i, M_{app}^i)}{2},
\end{equation}
where $M^i$ denotes the ground truth mask on $I_i$ and $\hat{M}^i$ represents the predicted one. In all of the following experiments, we report both \jf and \jt as metrics.

\textbf{Implementation Details.} Our method is build upon SAM2 framework, with MAE-pretrained~\cite{he2022masked} Hiera~\cite{bolya2023window, ryali2023hiera} serving as image encoders. We initialize SAM2 original modules with their official weights, firstly freeze other parameters, and train our transition detection module on IACC.3~\cite{IACC} and ClipShots~\cite{ClipShots} shot boundary detection datasets. In the following main training phase, we unfreeze all parameters and train the model for 30 epochs on YTVOS~\cite{2018YoutubeVOS} with TMA enabled. We set the number of sampling frames as 8 for the base-plus setting and 6 for the large. We enable focal, dice, iou, and CE losses in original SAM2, along with our proposed $\mathcal{L}_{box}$ and $\mathcal{L}_{exis}$. The weights of $\mathcal{L}_{box}$ and $\mathcal{L}_{exis}$ are both set as $0.5$. We employ AdamW as the optimizer, with the learning rate decaying from 5e-6 to 5e-7 during training. All experiments are conducted on 4 NVIDIA RTX-A6000 (48G) GPUs.

\begin{figure*}[t]
    \centering
    \includegraphics[width=\textwidth]{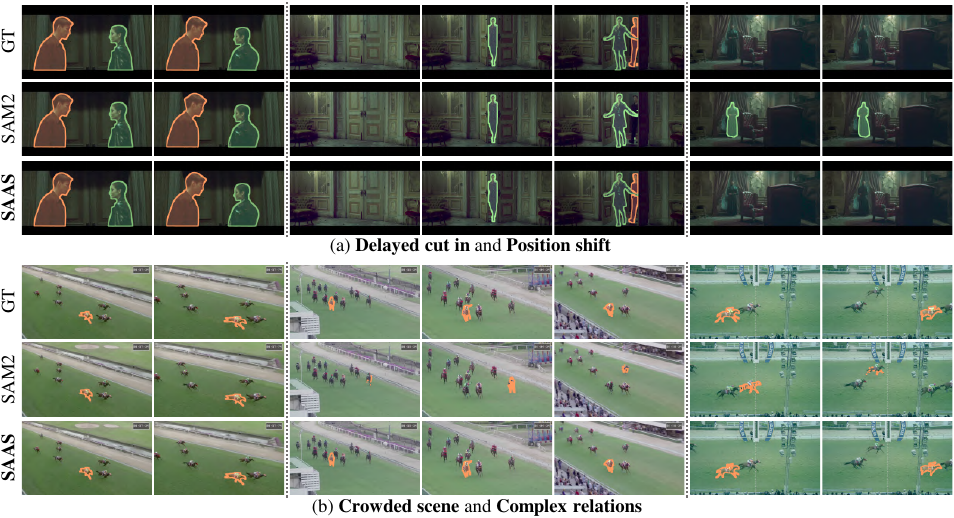}
    \caption{
    Qualitative comparison of some representative cases from \ourdataset between the \ourmethod and the SAM2 methods. (a) shows a case with a delayed cut in transition and an abrupt position shift of target objects. (b) demonstrates \ourmethod's better capacity in a crowded scene with complex relations. \ourmethod coherently segments the target object among ten similar objects.
    }
    \label{fig:qualitative}
\end{figure*}

\subsection{Main Results} \label{sec:main_results}
As shown in \Cref{tab:main_results}, we conduct exhaustive experiments on existing VOS methods~\cite{2022Xmem, cheng2023tracking, 2024Cutie, 2024SAM2} and our proposed SAAS on YouMVOS and Cut-VOS benchmarks. For SAM2 and \ourmethod, we test the base-plus setting and the large setting, respectively. The result shows that \ourmethod-B+ and \ourmethod-L outperform corresponding SAM2 methods and other existing VOS methods on two benchmarks across both \jt and \jf metrics, demonstrating its superiority. All reported data are calculated as the average of three runs.

From the table, we observe that training on YTVOS with TMA disabled (marked by $^\star$) results in a marginal improvement on YouMVOS (0.7\% \jf on Cutie and 1.3\% \jf on SAM2-B+). This strategy, however, suppressed methods' performance by 0.3\% to 0.9\% on Cut-VOS. The finding reveals that some videos from YouMVOS insufficiently represent MVOS difficulties, as they exhibit characteristics similar to conventional single-shot videos. In contrast, directly training on single-shot clips offers diminishing returns for Cut-VOS, which is specifically collected for MVOS.

The experimental result illustrates the effectiveness and robustness of the SAAS method. \ourmethod-B+ reaches 73.5\% \jf, 68.9\% \jt on YouMVOS(vs. 67.6\% and 63.7\%) and 60.7\% \jf, 53.1\% \jt on Cut-VOS(vs. 55.2\% and 47.2\%). Compared to SAM2-L, \ourmethod-L also attains consistent improvements of \jf (from 59.4\% to 62.0\%) and \jt (from 50.7\% to 54.0\% ). Notably, \ourmethod has virtually no degradation in inference speed due to efficient designs. Cutie+TMA method, compared to Cutie and Cutie$^\star$, reaches 69.6\% (vs. 68.4\%) and 53.5\% (vs. 52.3\%) \jf on two benchmarks, showing great generalization of TMA strategy. In the following ablation study, we offer more detailed data to further corroborate TMA and other modules' advantages. 

\subsection{Ablation Studies} \label{sec:ablation}
We analyze the validation of our modules via rigorous ablation studies, shown in~\Cref{tab:ablation_studies}. The ablation studies maintain the same implementation as the main experiments, employing the base-plus setting and uniformly tested on the Cut-VOS benchmark. In~\Cref{tab:ablation_studies}, we mainly study the effectiveness of different modules. Compared to baseline model I, the local memory bank $\mathcal{B}_{local}$ and TMA (model II and III) improve \jf by 2.4\% and 2.8\% respectively, while TMA plus TCH (V) achieves a 4.9\% \jf increase. For more ablation studies and hyperparameters analysis in detail, please refer to the appendix.
    
\subsection{Qualitative Results} \label{sec:qualitative}
\Cref{fig:qualitative} presents several representative visualized examples and corresponding segmentation results of SAM2 and \ourmethod models.
Case (a) shows a delayed cut in transition, one of the most difficult types, and a classical abrupt position shift of the target object, with a similar appearance distractor appearing at the same position.
SAM2 misses the target man (orange) when he reoccurs in shot 2, and incorrectly segments one another man with the same clothing (green) in shot 3, whereas our method successfully segments them.
In case (b), we highlight a crowded scene with complex relations between multiple similar objects.
SAM2 model struggles to match different instances correctly, leading to flickering predictions.
In contrast, by effectively capturing detail cues and establishing scene understanding, \ourmethod predicts high-quality masks for the object of interest consistently.
These examples demonstrate the superiority of our approach in complex multi-shot videos. A few more qualitative analyses are involved in the appendix.

\section{Conclusion}
\label{sec:conclusion}
We introduce \textbf{TMA}, a new training strategy that mitigates MVOS data sparsity by mimicking different transitions on single-shot datasets, and \textbf{SAAS}, a new MVOS method performing robust multi-shot segmentation capacity on complex edited videos. Meanwhile, we present a complex multi-shot benchmark, \textbf{Cut-VOS}, enabling evaluation and facilitating future research in MVOS. Extensive experiments demonstrate that our proposed strategy and method achieve state-of-the-art performance on MVOS benchmarks.

\textbf{Limitations}. Our method still struggles with extreme appearance changes of the target. For example, the same person with different clothing and hairstyles. The proposed TMA can't simulate this type effectively, and captured local cues may not help. This reflects one of the key challenges for MVOS: the model is required to both match unlike targets and distinguish similar distractors, demanding reducing the reliance on pure visual feature matching and requiring a stronger reasoning ability, which is to be further explored. 

\section*{Acknowledgments}
\footnotesize
This project was supported by the National Natural Science Foundation of China (NSFC) under Grant No. 62472104. 

\bigskip
\appendix

\section*{Appendix}

\section{Cut-VOS Benchmark}
\label{sec:a_dataset}

\subsection{Statistics}

The proposed benchmark, Cut-VOS, contains 100 high-quality videos, 174 objects of 11 different categories, 7965 frames, and 10.2K valid masks. Counting the number of shots for each target, Cut-VOS has 1131 shots in total. Counting the number of shots for each video, Cut-VOS has 648 shots, with an average of 6.5 shots per video. With an average length of 15.9 seconds of each video, the transition frequency is calculated as 0.346/s.

\begin{figure}[h]
    \centering
    \includegraphics[width=0.8\linewidth]{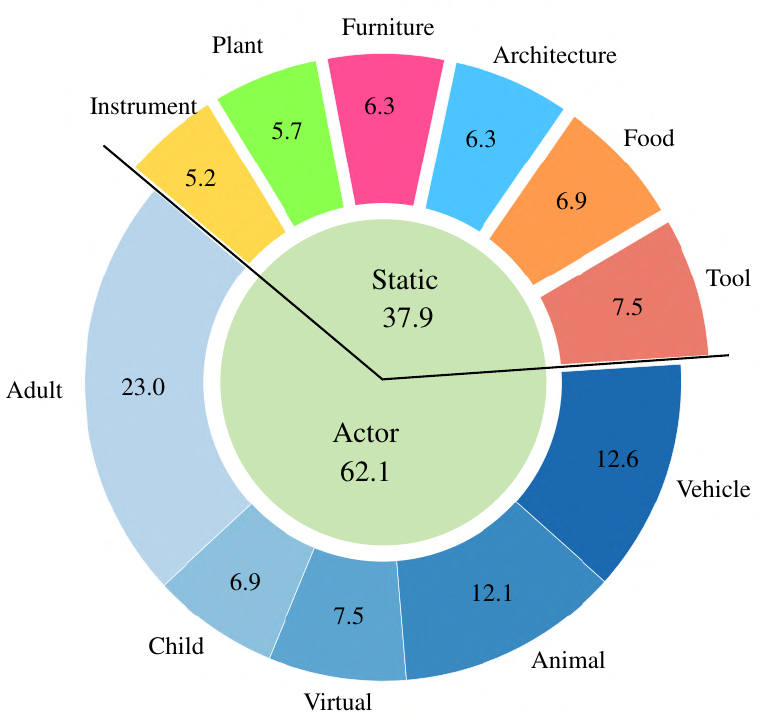}
    \caption{
    The object category distribution in Cut-VOS.
    }
    \label{fig:A_category}
\end{figure}

The distribution of object categories across 11 coarse categories is shown in ~\Cref{fig:A_category}. Cut-VOS benchmark has 62.1\% actors and 37.9\% static objects overall.

\begin{table}[h]
  \centering
  \normalsize
  \begin{tabular}{|l|l|l|l|l|}
  \myrule
      egg & flag & chair & flower & warplane \\
      \hline
      gun & lion & fruit & tomato & sculpture \\ 
      \hline 
      cup & sofa & horse & bottle & instrument \\
      \hline
      dog & meat & knife & insect & race car \\
      \hline 
      car & ball & motor & laptop & tree branch \\
      \hline 
      meat & bread & pizza & cheetah & other plants \\
      \hline 
      ship & adult & table & penguin & remote control \\
      \hline 
      stem & child & snake & building & other tool \\
      
  \myrule
  \end{tabular}
  \caption{All fine-grained categories involved in Cut-VOS.} 
  \label{tab:category}
\end{table}

In a finer classification manner, Cut-VOS contains 40 object categories. We show all contained categories in ~\Cref{tab:category}. Except for adult and child, which account for about 30\% of the total, the remaining categories show a relatively uniform distribution (3.2 instances per category on average).

\subsection{Transition Analysis}

In this section, we introduce different types of transitions classified by us in detail. ~\Cref{fig:A_transition} shows some representative cases picked from Cut-VOS. Except for the \textit{cut in} case we additionally show the conditional frame; we mainly show two preceding frames before the transition and two subsequent frames after the transition. With these visualization cases, we further explain the definition of these types:

\begin{figure}[h]
    \centering
    \includegraphics[width=\linewidth]{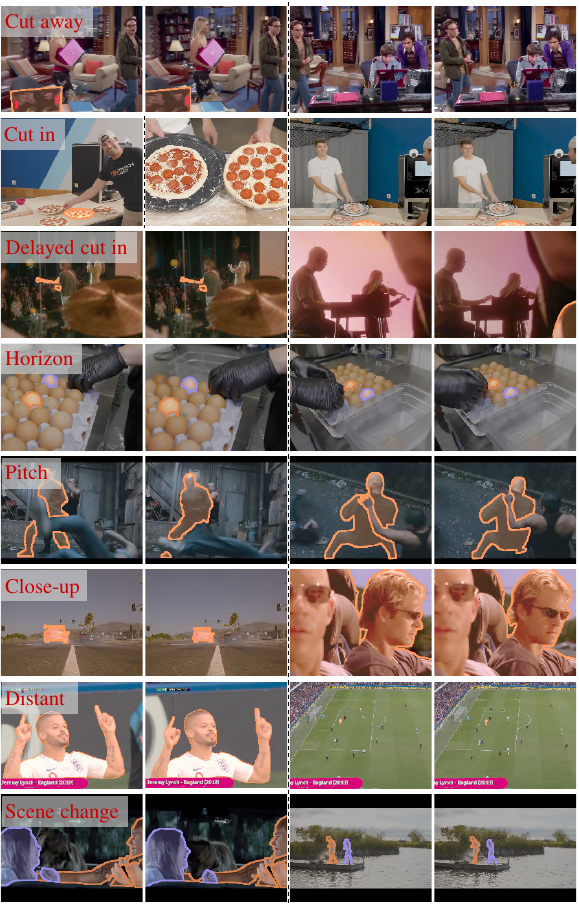}
    \caption{
    Cases of different types of transitions.
    }
    \label{fig:A_transition}
\end{figure}

\begin{enumerate}
    \item \textit{Cut away}. The target is present in the previous frame, but disappears in the next frame.
    \item \textit{Cut in}. The target isn't present in the previous frame, but reoccurs in the next frame.
    \item \textit{Delayed cut in}. The target may or may not be present in the previous frame. The target isn't present in the first frame of the next shot, but reoccurs in the shot with the movements of the camera view or the object itself.
    \item \textit{Horizon Transformation}. The camera rotates horizontally to reveal different aspects of the target.
    \item \textit{Pitch Transformation}. The camera's pitch angle changes to reveal the top or bottom surface of the target.
    \item \textit{Close-up view}. The viewpoint zooms in on the target, typically changing from the entire object to a partial.
    \item \textit{Distant View}. The viewpoint zooms out on the target, usually changing from a part of the object to the whole.
    \item \textit{Scene change}. Abrupt changes in time or space. The background would be completely different, sometimes with great changes in targets' appearance as well.
    \item \textit{Insignificance}. A cut between two similar shots (like a short frame decimation).
\end{enumerate}

\begin{algorithm}[t!]
\caption{\small{Transition Mimicking Data Augmentation}}
\label{alg:transition-mimic}
\definecolor{comment}{rgb}{0.2,0.2,0.6}
\definecolor{funcname}{rgb}{0.43,0.8,0.8}
\definecolor{variable}{rgb}{0.86,0.08,0.24}

\lstdefinelanguage{MyPython}{
    language=Python,
    morekeywords={[2] TMA, Affine_S, Affine_M, Hflip, CopyFg, GTranslation},
    alsoletter={_,<,>},
    morekeywords={[3] p_trans, p_once, p_cut, p_hflip, p_copy, p_same},
    deletekeywords={shape, range},
}
\lstset{
    backgroundcolor=\color{white},
    basicstyle=\fontsize{7.5pt}{7.5pt}\ttfamily\selectfont,
    columns=fullflexible,
    breaklines=true,
    captionpos=b,
    commentstyle=\fontsize{7.5pt}{7.5pt}\color{comment},
    keywordstyle=\fontsize{7.5pt}{7.5pt}\color{funcname},
    keywordstyle={[2]\bfseries}, 
    keywordstyle={[3]\bfseries\color{variable}}, 
    numbers=left,
    numbersep=4.5pt,
    xleftmargin=1.5em  
}

\begin{lstlisting}[language=MyPython]
def Affine_M(frame, mask):
    affine = RandomAffine(
        ... ) # a moderate affine
    return affine(frame), affine(mask)
def Affine_S(frame, mask):
    affine = RandomAffine(
        ... ) # a strong affine
    return affine(frame), affine(mask)
def Hflip(frame, mask):
    return frame[:,:,::-1], mask[:,:,::-1]
def CopyFg(s_frame, frame, mask):
    s_frame[mask>0] = frame[mask>0]
    return s_frame
def GTranslation(frame, mask, i, t):
    tl = RandomTranslation(1, 1)
    tl = trans * (t - i) / t
    return tl(frame), tl(mask)
def TMA(frames, masks, video_lists):
    # input format:[T,H,W,3],[T,H,W],VOSDataset
    if rand() > p_trans:
        return frames, masks
    if_once = rand() < p_once
    s = T // 2 if if_once else T // 3
    e = T if if_once else T // 3 * 2 + 1
    if rand() < p_cut:
        if_same_video = rand() < p_same
        if_copy = rand() < p_copy
        if if_same_video:
            aug_frames, aug_masks = 
                current_video.get_sample()
        # More possible to sample further segments
        else:
            aug_frames = video_lists.get_sample()
            aug_masks = np.zeros((T, H, W))
        aug_frames, aug_masks = Affine_M(
            aug_frames, aug_masks)
        for i in range(s, e):
            if not if_same_video and if_copy:
                frames[i], masks[i] = GTranslation(
                    frames[i], masks[i], i-s, T//2)
                aug_frames[i] = CopyFg(
                    aug_frames[i], frames[i], masks[i])
                aug_masks[i] = masks[i].copy()
            frames[i] = aug_frames[i]
            masks[i] = aug_masks[i]
    else:
        for i in range(s, e):
            frames[i], masks[i] = Affine_S(
                frames[i], masks[i])
    if rand() < p_hflip:
        for i in range(s, e):
            frames[i], masks[i] = Hflip(
                frames[i], masks[i])
    return frames, masks
    
\end{lstlisting}
\end{algorithm}

All types of transitions can be further classified into two categories: \textit{Cut away, cut in, and delayed cut in} as presence transitions, and the rest as view transitions. We allow the coexistence of one presence transition and one view transition when the target object reoccurs in the video, since the model is required to address the challenges arising from both types of transitions. It's difficult to conclusively determine which factor plays the predominant role in the model's failure. Also, multiple different view transitions may occur simultaneously (a 180-degree \textit{horizon transformation} and a 30-degree \textit{pitch transformation}, for example). We only marked the predominant one type (\textit{horizon}) for these transitions.

We also further explain the accuracy computation experiments based on the definitions of these transition types, mentioned in Section 4 of the main paper. The experiment is conducted on the SAM2-B+ model. For a given transition, we mark it as a correct segmentation if the IoU between the predicted masks and ground truth masks exceeds 0.5 in both the preceding frames before the transition and the subsequent frames after the transition (for delayed cut in, add the first frame where the target reoccurs as well). 

\subsection{Licences}

We will release our work under permissive open licences, including the Cut-VOS benchmark (CC by 4.0), the SAAS code(including pre-processing and evaluation codes), and main checkpoints(Apache 2.0).

\begin{table}[t]
  \centering
  \small
  \setlength{\tabcolsep}{4.2pt}
  \begin{tabular}{l|cccccc|cc}
  \myrule
      \multirow{2}{*}{ID} & \multicolumn{6}{|c|}{Hyperparameters} & \multirow{2}{*}{\jf} & \multirow{2}{*}{\jt} \\
      & $v_1$ & $v_2$ & $v_3$ & $v_4$ & $v_5$ & $v_6$ & & \\
      \hline
      I & 0.80 & 0.25 & 0.70 & 0.30 & 0.75 & 0.70 & 58.4 & 50.1 \\
      II & 0.75 & 0.30 & 0.70 & 0.40 & 0.75 & 0.60 & 58.8 & 50.2  \\
      III & 0.40 & 0.75 & 0.50 & 0.70 & 0.30 & 0.25 & 58.9 & 48.3  \\
      IV & 0.50 & 0.70 & 0.60 & 0.60 & 0.45 & 0.30 & 58.6 & 48.9 \\
      V & 0.50 & 0.50 & 0.60 & 0.50 & 0.50 & 0.45 & \underline{59.2}  & \underline{50.7}  \\
      VI & 0.60 & 0.60 & 0.50 & 0.40 & 0.75 & 0.55 & \textbf{59.5} & \textbf{51.6} \\

  \myrule
  \end{tabular}
  \caption{Experiments on the impact of different probability settings in TMA. The shortened variables $v_1$ to $v_6$ represent $p_{trans}$, $p_{once}$, $p_{cut}$, $p_{same}$, $p_{copy}$, $p_{hflip}$ respectively.} 
  \label{tab:tma_control}
\end{table}

\begin{table}[t!]
  \centering
  \small
  \setlength\tabcolsep{7.5pt}
  \renewcommand\arraystretch{1.1}
  \begin{tabular}{l|ccc|cc}
  \myrule
      ID & Aggregator & $Q_i$ & $\mathcal{B}_{scene}$ & \jf & \jt \\ 
      \hline
      I   & Linear &  \xmarkg  & \usym{2713} & 59.2 & 50.1  \\   
      II        & Convolution & \xmarkg & \usym{2713} & 58.9 & 50.2  \\
      III       & Convolution & \usym{2713} & \usym{2713} & \underline{59.9} & 50.9  \\
      IV      & Cross-attn & \xmarkg & \xmarkg & 58.2 & 51.2 \\
      V      & Cross-attn & \usym{2713} & \xmarkg & 59.6 & 51.4  \\
      VI     & Cross-attn & \xmarkg & \usym{2713} & 59.8 & \underline{51.7}  \\
      VII    & Cross-attn & \usym{2713} & \usym{2713} & \textbf{60.6} & \textbf{52.9}  \\
  \myrule
  \end{tabular}
  \caption{The ablation study on the design of TCH.}
  \label{tab:TCH_module}
\end{table}

\begin{enumerate}
    \item \textit{Do you have reason to believe the annotations in this dataset may change over time? Do you plan to update your dataset?} No.
    \item \textit{Are there any conditions or definitions that, if changed, could impact the utility of your dataset?} No.
    \item \textit{Will you attempt to track, impose limitations on, or otherwise influence how your dataset is used? If so, how?} The Cut-VOS benchmark would be released under a permissive CC by 4.0 licence.
    \item \textit{Were annotators informed about how the data is externalized? If changes to the dataset are made, will they be informed?} No.
    \item \textit{Is there a process by which annotators can later choose to withdraw their data from the dataset? If so, please detail.} No. 
\end{enumerate}

\section{Code}
\label{sec:a_code}

In this section, we mainly discuss our proposed data augmentation strategy and multi-shot segmentation method in detail. The main content includes: further explanatory notes on selected algorithms, comprehensive comparative experiments on hyperparameter configurations, sensitivity analysis of relevant parameters, \etc. Unless otherwise specified, all experiments in this section are conducted on Cut-VOS.

\subsection{Transition Mimicking Data Augmentation}

Firstly, we further describe the TMA algorithm in detail, including the introduction of all involved random variables used to control different transition patterns and how these patterns are actually generated. Overall, the workflow of the TMA algorithm is shown in 
~\Cref{alg:transition-mimic}.

\begin{table}[t]
  \centering
  \small
  \setlength\tabcolsep{6.5pt}
  \renewcommand\arraystretch{1.1}
  \begin{tabular}{l|cc|ccc}
  \myrule
      ID & $N_{enc}$ & $N_{dec}$ & \#Parameters(M) & \jf & \jt  \\ 
      \hline
      I & 1 & 1 & 91.4 & 58.1 & 49.8 \\
      II & 1 & 2 & 91.4 & 58.4 & 50.1 \\
      III & 2 & 1 & 92.4 & 59.0 & 50.6 \\
      IV & 2 & 2 & 92.5 & \textbf{59.4} & \textbf{51.2}  \\
      V & 3 & 3 & 93.6 & 59.1 & 50.6 \\
      VI & 4 & 4 & 94.8 & \underline{59.3} & \underline{50.8} \\   
  \myrule
  \end{tabular}
  \caption{The impact of the number of attention layers.} 
  \label{tab:layer_control}
\end{table}

\begin{table}[t!]
  \centering
  \small
  \setlength\tabcolsep{7pt}
  \renewcommand\arraystretch{1.1}
  \begin{tabular}{l|cc|cccc}
  \myrule
      ID & $\mathcal{L}_{exis}$ & $\mathcal{L}_{box}$ & $\mathcal{J}$ & $\mathcal{F}$ & \jf & \jt  \\ 
      \hline
      I & \ \ \ 0 & \ \ \ 0 & 57.3 & 60.7 & 59.0 & 49.0 \\
      II & 0.1 & 0.1 & 57.8 & 60.8 & 59.3 & 50.4 \\
      III & 0.5 & \ \ \ 0 & 57.7 & 60.8 & 59.2 & 50.2 \\
      IV & \ \ \ 0 & 0.5 & 58.1 & 60.5 & 59.3 & 49.8 \\
      V & 0.5 & 0.5 & \textbf{58.3} & \textbf{61.0} & \textbf{59.6} & \textbf{51.2} \\
      VI & \ \ \ 1 & \ \ \ 1 & \underline{58.3} & 60.9 & \underline{59.6} & \underline{50.8} \\ 
      VII & \ \ \ 4 & \ \ \ 4 & 57.9 & \underline{60.9} & 59.4 & 50.5 \\  
  \myrule
  \end{tabular}
  \caption{Comparison of model performance under different loss weights of $\mathcal{L}_{exis}$ and $\mathcal{L}_{box}$.} 
  \label{tab:loss_control}
\end{table}

The probability options are marked in red in the algorithm, involving $p_{trans}$, $p_{once}$, $p_{cut}$, $p_{sam}$, $p_{copy}$, and $p_{hflip}$. They work together to control the augmented data distribution. To better set their values, we conduct an exhaustive comparison of experiments on the TMA with different probability options. We train SAAS with 6 different settings for 20 epochs. The final results are reported in ~\Cref{tab:tma_control}.
Among these settings, settings I and II represent more aggressive augmentation strategies, tending to perform more numerous and complex augmentations with a higher likelihood to generate combined transformations. In contrast, settings III and IV adopt a more conservative strategy with lower frequency and mild augmentations. Settings V and VI are moderate, generating different transitions in a more balanced manner.

The experimental result shows that the moderate TMA settings are most beneficial to train MVOS models, reaching higher \jf and \jt on Cut-VOS. The observation offers reliable guidance for our final decisions on hyperparameters.

\subsection{Transition Comprehension Module}

We first study the different designs of the aggregator involved in TCH and how $\mathcal{B}_{scene}$ and $Q_i$ influence the model's performance. The results are shown in~\Cref{tab:TCH_module}. A Comparison between methods I, II, and VI reveals the best performance of the cross-attention aggregator (59.8\% vs. 59.2\% and 58.9\%). More experiments based on cross-attention aggregator (IV, V, VI, VII) further clarify the roles of $Q_i$ and $\mathcal{B}_{scene}$ in TCH.

To explore the specific architecture of the encoder to extract the transition state and the decoder to utilize $Q_i$ to refine previous memories, we conduct more experiments on it. Based on a common multi-head vision transformer layer~\cite{dosovitskiy2020image, liu2021swin} with a RoPE positional encoding~\cite{su2024roformer}, as reported in the main paper, we further adjust the number of stacked transformer layers ($N_{enc}$ and $N_{dec}$) to study the best design of them.  
For each experiment, we keep the other hyperparameters the same and retrain the model on the same hardware for 20 epochs. The final results are reported in \Cref{tab:layer_control}.The result reveals that an insufficient number of layers limits the model's expressive power, while excessive layers may introduce difficulties in training and convergence. Experimental results demonstrate that setting both the $N_{enc}$ and $N_{dec}$ to 2 yields an optimal performance. This model architecture is adopted in the other experiments.

We also conduct experiments to validate the effectiveness of two auxiliary objectives, which complement the ablation study presented in the main paper. We adjust the weights of $\mathcal{L}_{exis}$ and $\mathcal{L}_{box}$ for different settings and train each model for 20 epochs. The experiments covered the process of adjusting weights from small to large and different relative proportions to explore the optimal values.
The results are shown in~\Cref{tab:loss_control}. When the weights of $\mathcal{L}_{exis}$ and $\mathcal{L}_{box}$ are both set as 0.5, the model achieves a best performance of 59.6\% \jf and 51.2\% \jt. Compared with not using auxiliary objectives (I), it achieved a \jt improvement of approximately 2.2\%, reflecting the effectiveness of auxiliary objectives. As their weights increase (VI, VII), we observed a gradual decline in performance, possibly due to the impact on the primary mask objective during learning. Therefore, we keep them both as 0.5 in the main experiments.

\subsection{Training-free Memory Refinements}

\begin{table}[t]
  \centering
  \small
  \setlength\tabcolsep{9.5pt}
  \renewcommand\arraystretch{1.1}
  \begin{tabular}{l|c|cccc}
  \myrule
      ID & Groups & $\mathcal{J}$ & $\mathcal{F}$ & \jf & \jt  \\ 
      \hline
      I & \ ~1 & 58.0 & 60.7 & 59.3 & 50.7 \\
      II & \ ~2 & \underline{58.3} & \underline{61.1} & \underline{59.7} & \underline{51.2} \\
      III & \ ~4 & \textbf{58.6} & \textbf{61.4} & \textbf{60.0} & \textbf{51.3} \\
      IV & \ ~6 & 57.9 & 60.7 & 59.3 & 50.6 \\
      V & \ ~8 & 57.9 & 60.7 & 59.3 & 50.6 \\
      VI & 10 & 57.8 & 60.6 & 59.2 & 50.6 \\
      
  \myrule
  \end{tabular}
  \caption{The experimental results on the same checkpoint with different numbers of partition groups in the LMM.}
  \label{tab:group_table}
\end{table}
\begin{table}[t]
  \centering
  \small
  \begin{tabular}{l|lc|cccc}
  \myrule
      ID & Setting & $\tau_p$(\%) & $\mathcal{J}$ & $\mathcal{F}$ & \jf & \jt  \\ 
      \hline
      I & No Limit & \ \ 0.0 & \underline{58.6} & \underline{61.3} & \underline{59.9} & \textbf{51.5} \\
      II & Loose & \ ~1.0 & 58.4 & 61.1 & 59.8 & 51.2 \\
      III & Moderate & \ \ 2.5 & \textbf{58.6} & \textbf{61.4} & \textbf{60.0} & \underline{51.3} \\
      IV & Strict & \ \ 7.5 & 58.1 & 60.9 & 59.5 & 51.3 \\
      V & Extreme & 15.0 & 58.0 & 60.8 & 59.4 & 51.0 \\
      
  \myrule
  \end{tabular}
  \caption{SAAS's performance under different settings of $\tau_p$, tested on the same checkpoint.} 
  \label{tab:tau_table}
\end{table}
\begin{table*}[t!]
  \centering
  \small
  \setlength\tabcolsep{8pt}
  \begin{tabular}{lcccccccccc}
  \myrule
  \multirow{2}{*}{Method} & \multirow{2}{*}{Venue} & \multirow{2}{*}{FPS} & \multicolumn{4}{c}{YouMVOS$^\dagger$} & \multicolumn{4}{c}{Cut-VOS} \\
  \cmidrule(lr){4-7} \cmidrule(lr){8-11}
        &    &    & $\mathcal{J}$ & $\mathcal{F}$ & \jf & \jt & $\mathcal{J}$ & $\mathcal{F}$ & \jf & \jt \\
  \hline
    SAM2~\cite{2024SAM2}    & ICLR'25   &  \textbf{22} & 68.6  & 69.0  & 68.8  & 64.2  & 54.1  & 56.1  & 55.1  & 46.9  \\
    SAMURAI~\cite{yang2024samurai}    & Preprint'24 & 18 & 68.2 & 68.8 & 68.5 & 62.5 & 55.0  & 58.2  & 56.6  & 47.7      \\
    SAM2LONG~\cite{2024sam2long}   & ICCV'25   & 11 & 70.0 & 70.7 & 70.4 & 65.7  & 55.0  & 57.9  & 56.5  & 48.5      \\
    DAM4SAM (Videnovic et al. 2025)  & CVPR'25   & 17 & \underline{70.5} & \underline{71.6} & \underline{71.1} & \underline{65.9} & \underline{56.2}  & \underline{58.9}  & \underline{57.6}  & \underline{48.6}     \\
    \textbf{SAAS (Ours)}  & AAAI'26   & \underline{21} & \textbf{73.4} & \textbf{73.7}  & \textbf{73.5}  & \textbf{68.9}  & \textbf{59.4}  & \textbf{61.9} & \textbf{60.7} & \textbf{53.1} \\ 
  \myrule
  \end{tabular}
  \caption{Experiment results on more vision methods, including SAMURAI, SAM2LONGA, and DAM4SAM. They are all built upon the SAM2 model in a training-free manner.}
  \label{tab:other_methods}
\end{table*}
\begin{table*}[t!]
  \centering
  \small
  \setlength\tabcolsep{8.6pt}
  \begin{tabular}{lcccccccccccc}
  \myrule
    \multirow{2}{*}{Method} & \multicolumn{3}{c}{DAVIS2017-val} & \multicolumn{3}{c}{MOSE}
    &  \multicolumn{5}{c}{LVOSv2}  &  \multicolumn{1}{c}{YoutubeVOS} \\
    \cmidrule(lr){2-4} \cmidrule(lr){5-7} \cmidrule(lr){8-12} \cmidrule(lr){13-13}
      & $\mathcal{J}$ & $\mathcal{F}$ & \jf & $\mathcal{J}$ & $\mathcal{F}$ & \jf
      & $\mathcal{J}_s$ & $\mathcal{F}_s$ & $\mathcal{J}_u$ & $\mathcal{F}_u$ & \jf 
      & $\mathcal{G}$ \\
    \hline
    SAM2-B+ & 86.8 & 93.1 & 90.0 & 69.4 & \underline{77.4} & 73.4 & 80.1 & 87.0 & 78.3 & 85.1 & 82.6 & 88.2 \\
    SAAS-B+ & \underline{87.3} & 92.8 & 90.0 & 69.0 & 76.9 & 73.0 & 79.3 & 86.1 & \underline{80.3} & \underline{87.4} & 83.3 & 88.6 \\
    \hline
    SAM2-L & 86.9 & \textbf{93.4} & \underline{90.2} & \textbf{70.3} & \textbf{78.2} & \textbf{74.2} & \textbf{80.8} & \textbf{87.4} & 80.2 & 87.3 & \underline{83.9} & \underline{88.8} \\
    SAAS-L & \textbf{87.5} & \underline{93.1} & \textbf{90.3} & \underline{70.1} & 77.2 & \underline{73.6} & \underline{80.5} & \underline{87.3} & \textbf{81.6} & \textbf{89.0} & \textbf{84.6} & \textbf{89.2} \\

  \myrule
  \end{tabular}
  \caption{Experiment results on previous VOS datasets. We test the SAM2 and SAAS methods in a zero-shot setting.}
  \label{tab:VOS_experiments}
\end{table*}
\begin{table}[t]
  \centering
  \small
  \setlength\tabcolsep{8.6pt}
  \renewcommand\arraystretch{1.1}
  \begin{tabular}{lc|cccc}
  \myrule
    Method & Oracle & $\mathcal{J}$ & $\mathcal{F}$ & \jf & \jt \\
    \hline
    SAM2 & \xmarkg & 54.0 & 56.4 & 55.2 & 47.2 \\
    SAM2 & \usym{2713} & 91.1 & 95.1 & 93.1 & 97.0 \\
    SAAS & \xmarkg & 59.7 & 62.2 & 60.7 & 53.1 \\
    SAAS & \usym{2713} & 91.3 & 95.4 & 93.3 & 97.2 \\
  \myrule
  \end{tabular}
  \caption{Cross-shot oracle experiment on Cut-VOS.} 
  \label{tab:oracle_1}
\end{table}

The memory refinements introduced in the paper include a local memory bank to store local, fine-grained features and a scene memory bank used in TCH to build a basic understanding of the scene. These training-free refinements consistently enhance the performance, as shown in the ablation study in the main paper. In this section, we mainly focus on the technical details in the local memory bank $\mathcal{B}_{local}$.

We first explore the impact of the number of partition groups in the process of local memory bank extraction. Intuitively, too few groups may fail to segment the object into independent sub-regions, while too many groups could compromise semantic properties. We change the number of groups gradually, from 1 to 10, and benchmark the model performance on \textbf{the same} retrained checkpoint for a fair comparison. The results are shown in~\Cref{tab:group_table}.

As the number of groups increases, the model performance first improves, then declines, peaking at 4 groups (60.0\% \jf and 51.3\% \jt), confirming the hypothesis that a moderate number of groups exerts a positive influence on model performance. Thus, the partition groups are set as 4 in other experiments.
Another variable to be set is the proportion threshold $\tau_{p}$, which we use to control the construction of LMM to prevent over-partitioning on small targets. Thereby, its value may influence the quality of captured fine-grained features. We conduct a comparison experiment to study this via testing the SAAS segmentation results on the Cut-VOS under 5 settings with different values of $\tau_p$: no limitation~(0\%), loose~(1\%), moderate~(2.5\%), strict~(7.5\%), and extreme strict~(15\%). The results are illustrated in~\Cref{tab:tau_table}.

Overall, the model seems not really sensitive to $\tau_p$ of small values~($\le 2.5\%$). However, when $\tau_p$ gets larger, our proposed local memory bank no longer fulfills its intended function, leading to a plainly visible degradation. The result indicates that further partitioning the small objects into pixel pieces won't lead to really downside. But we are still willing to set it as a moderate value, \eg. 2.5\%, considering the robustness and reasonability.

\subsection{SAAS Hyperparameters Selection}

Drawing from the comprehensive results of the above comparison experiments, we establish a well-defined hyperparameter configuration for the SAAS method, supporting the main results presented in the paper. We set $p_{trans}$, $p_{once}$, $p_{cut}$, $p_{same}$, $p_{copy}$, $p_{hflip}$ as $0.60$, $0.60$, $0.70$, $0.40$, $0.75$, $0.55$ in TMA respectively, adopting a relatively balanced strategy. For the transition comprehension module, we set the number of both encoder layers and decoder layers as $2$. The structure of the aggregator is decided as cross-attention layers, following the result of ablation studies in the main paper. Two new auxiliary objectives are both enabled, with a weight of $0.05$. The local memory bank $\mathcal{M}_{local}$ is constructed with local detail features from 4 sub-regions, only if the proportion of the ground truth mask exceeds 2.5\% in the conditional frame. The entire configuration works well in exhaustive experiments, outperforming the baseline significantly and achieving 74.2\% \jf on YouMVOS and 62.5\% \jf on Cut-VOS. We believe this content will help reproduce our method on other devices.
\section{Experiments}
\label{sec:a_experiments}

\begin{figure*}[t!]
    \centering
    \includegraphics[width=0.99\linewidth]{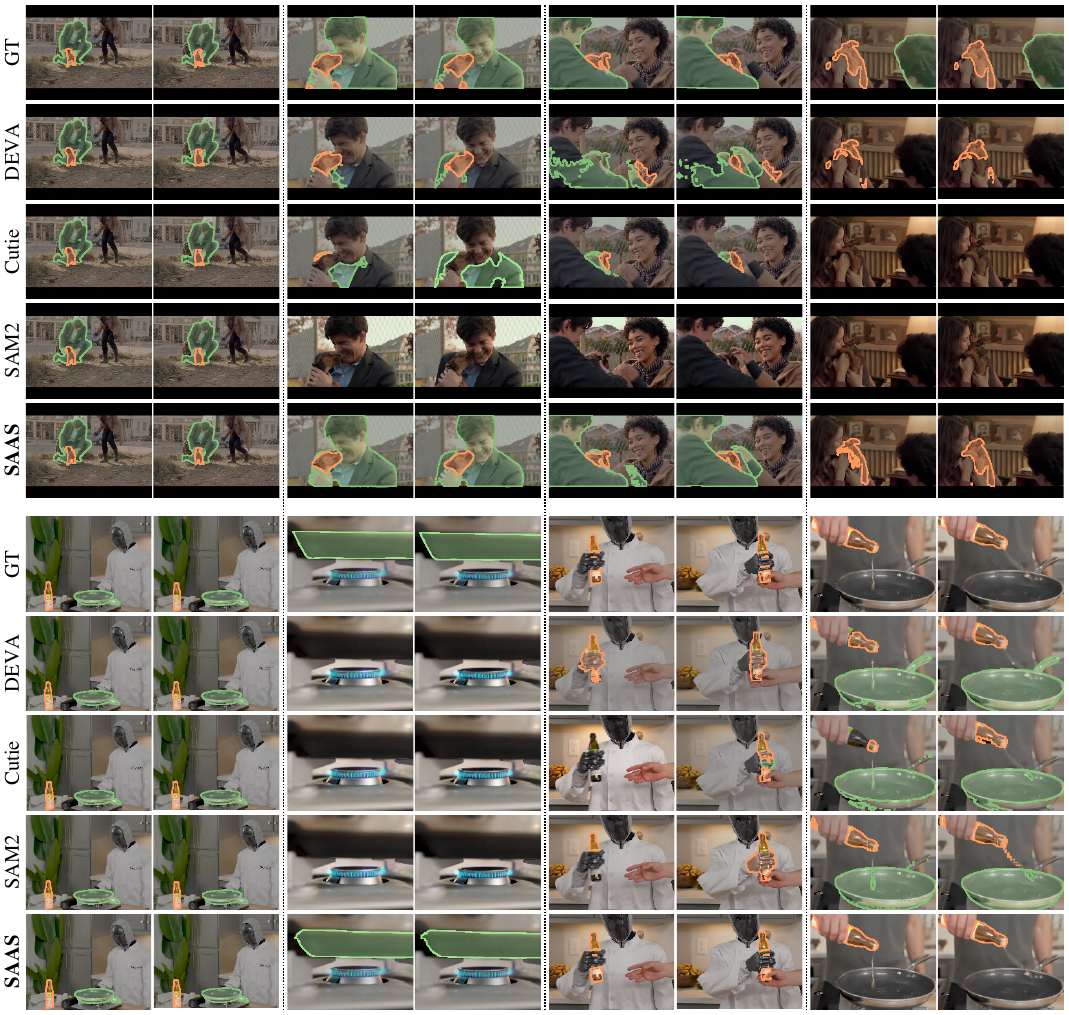}
    \caption{
    More qualitative experiment results, including a video containing many shot transitions of different types, and a case to segment static objects. SAAS performs robust segmentation capacity on these complex videos.
    }
    \vspace{-3mm}
    \label{fig:more_qualitative}
\end{figure*}

\subsection{Comparison to More Vision Methods}

We benchmark some most recent approaches that are building upon SAM2 as well, \eg. SAMURAI~\cite{yang2024samurai}, SAM2LONG~\cite{2024sam2long}, and DAM4SAM~\cite{2025distractor}. Since these methods are designed in a training-free manner, we train another SAM2-B+ checkpoint under the same experiment setting and benchmark these methods with the only checkpoint for a fair comparison. The results are shown in~\Cref{tab:other_methods}. The results reveal that despite marginal improvements brought by these methods, they still struggle with complex multi-shot videos. Also, they significantly lag behind our proposed SAAS, which is specifically introduced for MVOS, across both \jf and \jt metrics. The second-best method, DAM4SAM, improves \jf from 55.1\% to 57.6\% compared to the baseline, but shows a 3.1\% \jf and 4.5\% \jt gap to SAAS. The experiments highlight the SAAS's superiority on MVOS compared to existing vision methods.

\subsection{Performance on Previous VOS Datasets}
\label{sec:VOS_results}

We also report the performance of SAAS on some previous single-shot VOS datasets, including DAVIS2017~\cite{pont20172017}, MOSE~\cite{2023MOSE}, LVOSv2~\cite{hong2024lvos}, and YoutubeVOS~\cite{2018YoutubeVOS}, compared to the baseline SAM2 model. The main results are shown in~\Cref{tab:VOS_experiments}. For each dataset, we adopt their official repositories or websites for evaluation for a fair comparison. The experiment follows a zero-shot setting, where we directly generalize the official SAM2 model checkpoints and our checkpoints from the main experiments to these datasets. 

\begin{figure*}[t!]
    \centering
    \includegraphics[width=0.998\linewidth]{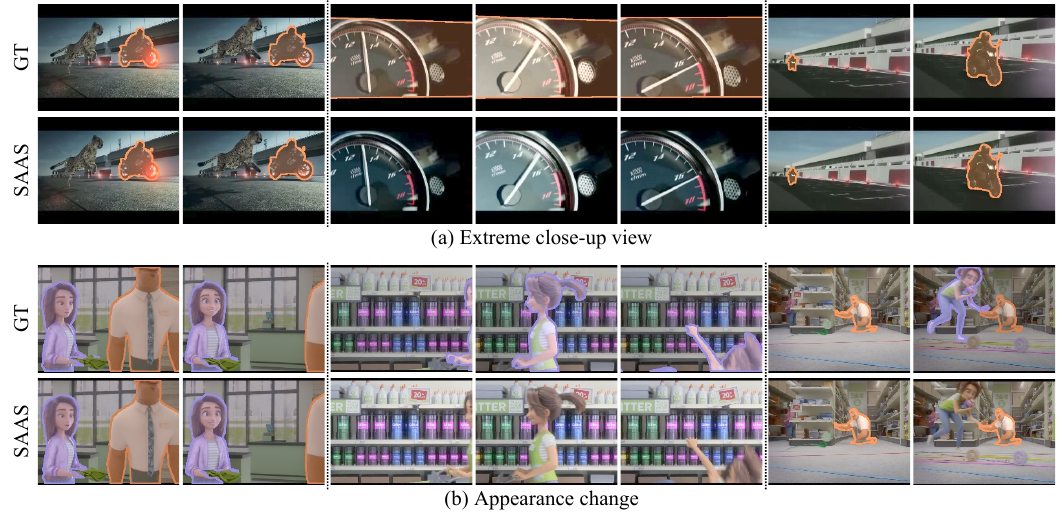}
    \caption{
    Some visualized failure cases of the SAAS method on the Cut-VOS benchmark. Case (a) shows an extreme close-up view that suddenly zooms in on the dashboard of a motor. Case (b) requires the model to match the same person with different clothing and hairstyles. Our SAAS method still has difficulties correctly segmenting these cases, which require a stronger reasoning ability.
    }
    \vspace{-2mm}
    \label{fig:failure_case}
\end{figure*}

The experimental result demonstrates that SAAS achieves a similar performance on single-shot videos compared to the baseline model. On DAVIS2017-val, SAM2 and SAAS exhibit nearly identical performance. SAAS achieves a  0.5\% performance improvement on LVOSv2 and YoutubeVOS, while showing a marginal degradation on the complex VOS dataset MOSE. 

We consider the comparable performance to the baseline to be expected, as the optimizations in our method rely on the shot transition detection, without incorporating additional designs for single-shot videos. The observation that the improvement on LVOSv2 is mainly brought by unseen objects may demonstrate that our method achieves more accurate detection of object disappearance and reappearance.
For the decrease of \jf on MOSE, our SAAS primarily lags behind SAM2 in segmentation contour quality ($\mathcal{F}$), highly likely caused by serious occlusion phenomena in MOSE. The TMA strategy lacks a specialized design for occlusion scenarios, directly placing objects on the top layer during replication. Additionally, occlusion instances are underrepresented in the training dataset~(YTVOS). The bias in the data distribution may cause the drop in $\mathcal{F}$.

\subsection{Oracle Experiments}

In this section, we supplement an oracle experiment which assumes the models possess a perfect cross-shot segmentation module. This means that the models can always segment the target object correctly in each shot segment, i.e., the ground truth mask is provided in the first frame of each shot. The result is shown in~\Cref{tab:oracle_1}. 

The \jt metrics haven't reached 100\% mainly due to the delayed cut in transition type, which requires not only the first ground truth mask of the shot. Also, some refinements made by the model bring minor disturbances. However, the \jf and \jt have reached a very high level overall. This reveals that the most challenging aspect of our proposed Cut-VOS benchmark lies in complex shot transitions, rather than coherent segmentation within the same shot. While existing methods can achieve a high-quality \textit{intra-shot} segmentation, they struggle with \textit{inter-shot} segmentation. Under the premise of disregarding shot transitions, the SAM2 model and SAAS method demonstrate comparable segmentation performance, which aligns with the experimental results in~\Cref{sec:VOS_results}.

\subsection{Qualitative Results}

We provide more qualitative experiment results on our Cut-VOS benchmark, as shown in~\Cref{fig:more_qualitative}, to further show the superiority of the proposed SAAS method compared to the existing VOS models. In the upper case, the video contains many transitions of different types, including \textit{close-up view}, \textit{horizon transformation}, \textit{scene change}, \textit{etc}. The SAM2 model misses the target objects at a very early time, while the SAAS method correctly tracks and segments them, despite some minor artifacts. In the down case, we mainly show a sample to track static objects. Though these static objects don't perform complex motion patterns and significant appearance changes, the SAM2 model struggles to correctly track them coherently and distinguish them among similar distractors in complex multi-shot videos. In contrast, SAAS exhibits improved segmentation performance.

\subsection{Limitations and Failure Cases}

In this section, we further investigate the latent limitations of the SAAS method. Upon careful examination of the generated mask, we observe that, despite enhanced object matching across shots, SAAS still relies on visual feature matching, which lacks robust long-range reasoning. This deficiency contributes to segmentation errors, especially in cases where the target object's appearance completely changes during shot transitions, as depicted in~\Cref{fig:failure_case}. Case (a) involves an extreme \textit{close-up view} transition, rapidly zooming in on a local sub-region of the object that was entirely unseen in the previous frames, rendering some of our advancement strategy ineffective. In such cases, humans rely on commonsense knowledge or inferences about filming intent for correct decision-making, a process that poses significant challenges for current methods. In case (b), we highlight a girl which have different clothing and hairstyles across the shots. Unluckily, our method tends to misidentify her as a separate individual, treating her as a distractor. Our proposed TMA strategy can not effectively simulate similar scenarios, which may be a key contributing factor to the observed phenomenon.

\bibliography{aaai2026}


\end{document}